\documentclass{article}
\usepackage{spconf}
\usepackage{amsmath,amssymb,amsfonts}
\usepackage{algorithmic}
\usepackage{graphicx}
\usepackage{textcomp}
\usepackage{xcolor}
\usepackage{hyperref}

\usepackage[eprint=false,isbn=false,url=false,style=ieee,maxnames=20,minnames=1]{biblatex}   
\AtEveryBibitem{
 \clearlist{address}
 \clearfield{date}
 \clearfield{eprint}
 \clearfield{isbn}
 \clearfield{issn}
 \clearlist{location}
 \clearfield{month}
 \clearfield{day}
 \clearfield{series}
 \clearfield{volume}
 \clearfield{number}
 \clearfield{pages}
 \clearfield{note}
 \clearlist{language}
 \ifentrytype{book}{}{
  \clearlist{publisher}
  \clearname{editor}
 }
}
\AtBeginBibliography{\footnotesize}
\begin{document}
\ninept

\title{Shift- and stretch-invariant non-negative matrix factorization with an application to brain tissue delineation in emission tomography data}

\name{
\begin{tabular}{@{}c@{}}
\textit{Anders S. Olsen}$^{1*}$, 
\textit{Miriam L. Navarro}$^{1,2}$, 
\textit{Claus Svarer}$^1$,
\textit{Jesper L. Hinrich}$^{3}$, \\
\textit{Morten Mørup}$^{3}$, 
\textit{Gitte M. Knudsen}$^{1,4}$
\end{tabular}} 

\address{
\small
$^{1}$\textit{Neurobiology Research Unit, Copenhagen University Hospital Rigshospitalet, Copenhagen, Denmark}
\\
\small
$^{2}$\textit{Department of Neuroscience, Faculty of Health and Medical Sciences, University of Copenhagen, Copenhagen, Denmark}
\\
\small
$^{3}$\textit{Department of Applied Mathematics and Computer Science, Technical University of Denmark, Kgs. Lyngby, Denmark}
\\
\small
$^{4}$\textit{Department of Clinical Medicine, Faculty of Health and Medical Sciences, University of Copenhagen, Copenhagen, Denmark}
\\
\small
$^*$\textit{Corresponding author: anders.stevnhoved.olsen@gmail.com}
}

\maketitle

\begin{abstract}

Dynamic neuroimaging data, such as emission tomography measurements of radiotracer transport in blood or cerebrospinal fluid, often exhibit diffusion-like properties. These introduce distance-dependent temporal delays, scale-differences, and stretching effects that limit the effectiveness of conventional linear modeling and decomposition methods. To address this, we present the shift- and stretch-invariant non-negative matrix factorization framework. Our approach estimates both integer and non-integer temporal shifts as well as temporal stretching, all implemented in the frequency domain, where shifts correspond to phase modifications, and where stretching is handled via zero-padding or truncation. The model is implemented in PyTorch (\href{https://github.com/anders-s-olsen/shiftstretchNMF}{https://github.com/anders-s-olsen/shiftstretchNMF}). We demonstrate on synthetic data and brain emission tomography data that the model is able to account for stretching to provide more detailed characterization of brain tissue structure. 

\end{abstract}
\section{Introduction}
\label{sec:intro}


Dynamic medical imaging, e.g., positron emission tomography (PET) or single-photon emission computed tomography (SPECT) enables tracking of radiotracer kinetics and physiological processes across space and time. These measurements can provide valuable insights into brain function including movement of cerebrospinal fluid (CSF) or binding patterns of neuroreceptors. 

However, accurate analysis of tracer-based imaging data is hindered by several challenges. Extracting regional signals typically requires aligning emission data with high-resolution anatomical images (MRI or CT). This registration step is error-prone due to subject motion and physiological variability between scans. Furthermore, partial volume effects blur tissue boundaries, degrading the precision of regional estimates \cite{marquis_partial_2023}. An alternative strategy is to model the latent composition of the emission data, bypassing anatomical information. By clustering or decomposing raw time–activity curves (TACs), one can identify distinct regions with unique kinetic signatures and thus mitigate alignment and resolution issues.

\begin{figure}[h]
    \centering
    \includegraphics[width=1\linewidth]{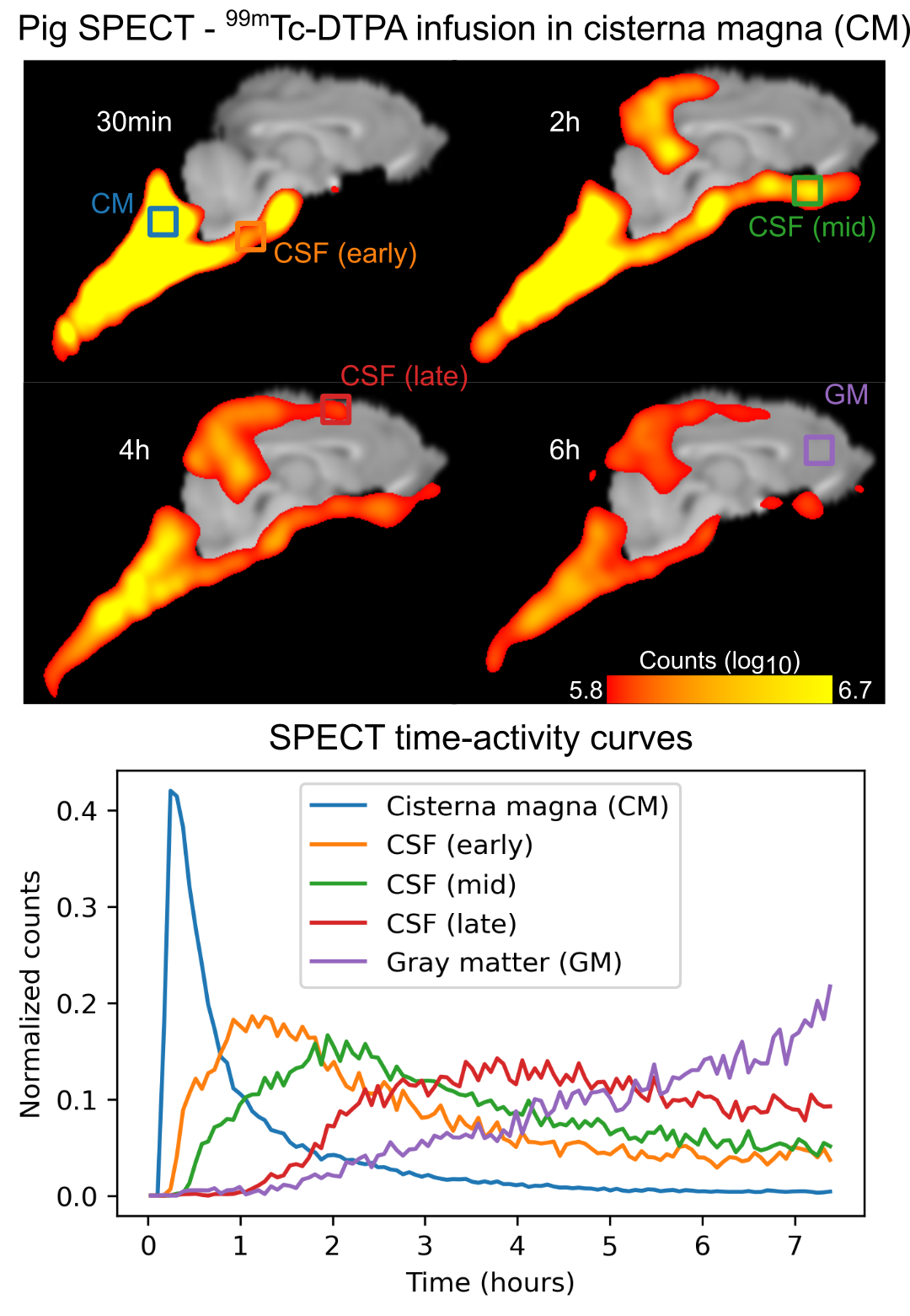}
    \caption{SPECT data from an example pig after $\textsuperscript{99m}$Tc-DTPA radiotracer injection in the cisterna magna (CM). Time-activity curves in the time-series plot - from selected voxels and timepoints indicated by squares - were normalized to unit norm for visualization.}
    \label{fig:spect}
\end{figure}

Various clustering and decomposition techniques have been applied to this task, including K-means, fuzzy C-means, Gaussian mixture models, and non-negative matrix factorization (NMF) \cite{acton_automatic_1999,ashburner_chapter_1996,boudraa_delineation_1996,jaakkola_segmentation_2023,kim_noninvasive_2001,lee_non-negative_2001}, see \cite{rainio_quantitative_2025} for a recent review. While effective in certain contexts, these approaches generally assume that components are fixed in time and therefore fail to account for delays or dispersion of TACs across regions, evident as relative shifts and stretching of TACs, which can result in misclassification or blurred regional boundaries. Fig.~\ref{fig:spect} shows example data, where a radiotracer is infused in the cisterna magna (CM) and subsequently disperses in the CSF and into the gray matter. This causes TACs to differ in onset and tracer movement speed, leading to delays and stretching even between areas of the same tissue type. 

To overcome this limitation, we propose a shift- and stretch-invariant extension of NMF. Building on shift-invariant NMF \cite{morup_shifted_2007}, which estimates integer (i.e., whole samples) and non-integer (sub-sample) temporal delays, our model additionally incorporates stretch-invariance to capture differences in the speed of tracer kinetics. This flexibility enables more accurate alignment of TACs across tissues exhibiting heterogeneous dynamics. We validate the method on synthetic data and animal brain SPECT experiments, and provide an open-source PyTorch toolbox (\href{https://github.com/anders-s-olsen/shiftstretchNMF}{https://github.com/anders-s-olsen/shiftstretchNMF}).

\section{Methods}
\label{sec:methods}

\subsection{Shift- and stretch-invariant non-negative matrix factorization}
Given a data set $\mathbf X\in\mathbb{R}^{P\times N}$, where $j=1,\ldots,P$ indexes channels (e.g., voxels in imaging data) and $t=1,\ldots,N$ indexes temporal samples, the standard bilinear factor analysis model is 
\begin{equation}
    x_{j}(t)\approx\sum_k a_{j,k}\,s_{k}(t),\label{eq:nmf}
\end{equation}
yielding a set of $K$ channel maps $\mathbf A\in\mathbb{R}^{P\times K}$ and temporal profiles $\mathbf S\in\mathbb{R}^{K\times N}$. Additional constraints may be imposed on the decomposition, such as statistical independence in independent component analysis \cite{bell_information-maximization_1995} or non-negativity in NMF \cite{lee_learning_1999}, the latter being particularly suitable when the data itself is non-negative. 

The shift-invariant factor analysis extends this framework by allowing temporal profile shifts (delays) for each channel \cite{harshman_shifted_2003}:
\begin{equation}
    x_{j}(t)\approx\sum_k a_{j,k}\,s_{k}\left(t-\tau_{j,k}\right), \label{eq:shiftnmf}
\end{equation}
with $\tau_{j,k}\in\mathbb{R}$ denoting the per-channel, per-component shift (temporal delay in TACs). Shift-invariant NMF estimation has been proposed using multiplicative updates \cite{morup_shifted_2007} and a time-frequency gradient method \cite{madsen_time_2009}. A different generalization of NMF is to account for stretch (dispersion) by expanding or compressing the temporal profiles (time-scaling):
\begin{equation}
    x_{j}(t)\approx\sum_k a_{j,k}\,s_{k}\left(t/r_{j,k}\right),\label{eq:stretchnmf}
\end{equation}

with $r_{j,k}\in\mathbb{R}_+$ denoting stretch ($r>1$) or compression ($r<1$). This idea was recently explored using spline interpolations in the temporal domain \cite{gu_stretched_2024}. 

Here, we propose combining both shift- and stretch-invariance in the same model such that for each channel $j$ and component $k$, both a shift $\tau_{j,k}$ and a stretch $r_{j,k}$ is learned:

\begin{equation}
    x_{j}(t)\approx\sum_k a_{j,k}\,s_{k}\left(\frac{t-\tau_{j,k}}{r_{j,k}}\right).\label{eq:shiftandstrechnmf}
\end{equation}

In the frequency domain, temporal shifts correspond to constant phase modulations and stretching to frequency scaling by the inverse time-scaling factor. Consequently, working in the frequency domain makes model estimation more computationally efficient:

\begin{equation}
    \tilde{x}_{j}(f)\approx \sum_k a_{j,k}\,\tilde{s}_{k}\left(r_{j,k} f\right)e^{-i2\pi \frac{f}{N}\tau_{j,k}},
\end{equation}

where $\tilde{\mathbf s}_k$ and $\tilde{\mathbf x}_j$ are one-sided discrete Fourier transforms (DFT), $i=\sqrt{-1}$, and $f=0,\ldots,N_{FFT}$. Of note, in the \textit{continuous} Fourier transform, the time-frequency scaling property requires amplitude modulation to satisfy Parseval's theorem. In the DFT, no closed-form analogue exists for non-integer $r$ \cite{talwalkar_time-frequency_2010}. Instead, we rescale the coefficients of truncated spectra/signals empirically (see Sec.~\ref{sec:stretch}). Phase shifts are frequency-independent, hence frequency scaling only applies to $\tilde{s}_k(f)$. 

In condensed form, the loss function to be optimized is

\begin{equation}
    \mathcal{L}=\frac{1}{N}\sum_{j=1}^P\bigg\lVert\mathbf w\odot\left(\tilde{\mathbf x}_j-\sum_k \tilde{\mathbf a}^{\left(\tau_{j,k}\right)}_{j,k}\odot\tilde{\mathbf s}^{\left(r_{j,k}\right)}_k\right)\bigg\rVert^2,
\end{equation}

where $\odot$ is element-wise multiplication, $\tilde{\mathbf a}_{j,k}^{\left(\tau_{j,k}\right)}=a_{j,k}\,e^{-i2\pi\frac{\mathbf f}{N}\tau_{j,k}}$ is a vector of size $N_{FFT}$ and $\tilde{\mathbf s}_k^{\left( r_{j,k}\right)}$ contains frequency-scaled component profiles. The vector $\mathbf w=[\frac{1}{\sqrt 2},1,\ldots,1,\frac{1}{\sqrt 2}]$ ensures correct weighting of DC and Nyquist terms for comparison between time-domain and one-sided frequency domain loss functions according to Parseval's identity, which establishes a direct relation between the time and frequency domain signal energy. 

\subsection{Estimation of shifts}

Non-linear optimization of the shift-parameter $\tau_{j,k}$ in previous studies using the Newton-Raphson method found that estimated shifts seldom exceeded one temporal index due to a large distance between minima in the parameter landscape \cite{morup_shift-invariant_2008,morup_shifted_2007}. Instead, shifts were proposed to be estimated first using the cross-correlation function for integer shifts and subsequently non-linear finetuning for obtaining non-integer shifts. For each component $k$, we compute the residual spectrum with all other sources projected out: 
\begin{equation}
    \tilde{g}_{j,k}(f)=\tilde{x}_{j}(f)-\sum_{k'\neq k}\tilde{a}_{j,k'}^{\left(\tau_{j,k'}\right)}(f)\,\tilde{s}_{k'}(f).
\end{equation}
The cross-spectrum is $\tilde{h}_{j,k}(f)=\tilde{g}_{j,k}(f)^{*}\,\tilde{s}_{k}(f)$, where $(\cdot)^*$ is the complex conjugate. The inverse DFT gives the cross-correlation function where the optimal delay is
\begin{equation}
    l_{j,k}=\arg \max_t h_{j,k}(t), \quad \tau_{j,k}=l_{j,k}-N,
\end{equation}
assuming zero-based indexing. We note that computing the cross-correlation function can be avoided by estimating the slope of the phase of the cross-spectrum $\angle\tilde{h}_{j,k}(f)$. However, this requires “unwrapping” the phase to avoid discontinuities at $\pm\pi$ and combined with the slope estimation, this turns out to actually be slower than the inverse DFT of the cross-spectrum. 

\subsection{Estimation of stretches}\label{sec:stretch}
Here we extend the above cross-correlation method to also simultaneously estimate stretches. We start by constructing a library of spectral profiles $\tilde{s}_k^{(b)}(f)$, $b=[-N_{FFT}/2+1,\ldots,0,\ldots,N_{FFT}/2-1]$ by truncating the spectrum ($b<0$) or zero-padding ($b>0$). To obtain the same $N_{FFT}$ for all profiles we computed the inverse DFT and zero-padded in the time-domain ($b<0$) or truncated the temporal representation ($b>0$) followed by the forward DFT. In the case of truncating in either domain, the retained coefficients were rescaled by the square root of the ratio of the original energy to the truncated energy to ensure the total energy was conserved. During optimization, the library is constructed for each update of $\mathbf S$. Of note, the integer index variable $b$ is related to the continuous frequency axis scaling parameter through $r=1+\frac{b}{N_{FFT}}$. 

Residuals are then computed as

\begin{equation}
    \tilde{g}_{j,k}(f)=\tilde{x}_{j}(f)-\sum_{k'\neq k}\tilde{a}_{j,k'}^{\left(\tau_{j,k'}\right)}(f)\,\tilde{s}^{\left(b_{j,k'}\right)}_{k'}(f),
\end{equation}

where $b_{j,k}$ indexes into the library of spectral profiles. The cross-spectrum is computed over $\tilde{h}^{(b)}_{j,k}(f)=\tilde{g}_{j,k}(f)^{*}\,\tilde{s}^{(b)}_{k}(f)$ such that for each channel $j$ and component $k$, it is now a 2D maximization problem over dimensions $b$ and $t$. 

\begin{equation}
    [l_{j,k},b_{j,k}]=\arg \max_{t,b} h_{j,k}^{(b)}(t), \quad \tau_{j,k}=l_{j,k}-N.
\end{equation}

\subsection{Estimation of $\mathbf A$ and $\mathbf S$}

Given the estimated shifts and stretches, an analytical estimate of the matrix $\mathbf A$ is given in closed form (extended from \cite{morup_shifted_2007} to also include stretching):

\begin{equation}
    a_{j,k} = \frac{h_{j,k}^{(b_{j,k})}(l_{j,k})}{\tilde{\mathbf s}_k^\mathrm{H}\tilde{\mathbf s}_k^{}},
\end{equation}

i.e., the value of the maximal cross correlation scaled by the component energy, where $(\cdot)^\mathrm{H}$ is the Hermitian transpose operator. Any negative elements in $\mathbf A$ are clipped to zero.

We propose non-linear optimization of $\mathbf S$ in PyTorch, which has repeatedly shown to produce comparable results to models estimated via analytically derived update rules for various non-convex unsupervised problems with appropriate initialization while being considerably faster \cite{olsen_uncovering_2024,olsen_coupled_2024}. Using this approach, we optimize $\bar{\mathbf S}$ (S-bar) unconstrained while the softplus function (which is differentiable everywhere) is used to impose element-wise non-negativity prior to calculating shifts, stretches, and loss, i.e., $s_{k,t}=\ln \left(1+e^{\bar{s}_{k,t}}\right)$. Non-negativity in the time-domain does not have a direct equivalent in the frequency domain, so $\bar{\mathbf{S}}$ is in the time-domain and the spectral library of stretches is established for each iteration. We used ADAM (learning rate $0.1$) in all experiments. The stopping criterion searched the latest 50 iterations, selected the lowest two losses, and stopped optimization if the relative change between the two losses was below $10^{-10}$ or if the loss only increased.

\subsection{Data}

We used experimental data from five pigs infused with $\textsuperscript{99m}$Tc-DTPA in the CM immediately followed by $N\approx100$ SPECT-acquisitions, each accumulating radiation counts over six minutes (see an example subject in Fig.~\ref{fig:spect}). These data were acquired to investigate radiotracer flow from CSF to the brain parenchyma to describe the glymphatic system in gyrated brains. The poor spatial resolution of SPECT data renders it particularly susceptible to partial volume effects \cite{marquis_partial_2023} and radiotracer movement is slow \cite{nilsson_circadian_1992}, making a shift- and stretch-invariant model particularly viable to extract whole-brain tissue compartments. Due to vast scale differences between tissue compartments, the data in each voxel was normalized to unit norm. To avoid inflating noise channels, only voxels with summed counts higher than $5\cdot 10^6 Bq/cc$ were included. Moreover, the spinal cord was excluded, totaling $P\approx20.000$ brain voxels. Representing shifts in the frequency domain assumes circular shifting, and to avoid late data entering as early data when shifted, we zero-padded the TACs by 20\% of $N$ in the time-domain. 

\subsection{Experiments}


Combined, we tested four models: (1) NMF, (2) Integer-shift NMF, (3) Non-integer-shift NMF (initialized from (2)), and (4) Shift-stretch NMF (initialized from (2)). For all models, we used K-shape (K-means with the cross-correlation distance \cite{paparrizos_k-shape_2016}) for initialization, which we found to perform marginally better than other typical NMF-initializers (not shown). This procedure only initialized $\mathbf S$ while an initial guess for $\mathbf A$ was entered using a least-squares solution. Models (1) and (3) also optimized $\mathbf A$ non-linearly in PyTorch, since these models had no access to the cross-correlation function.

\begin{figure}
    \centering
    \includegraphics[width=1\linewidth]{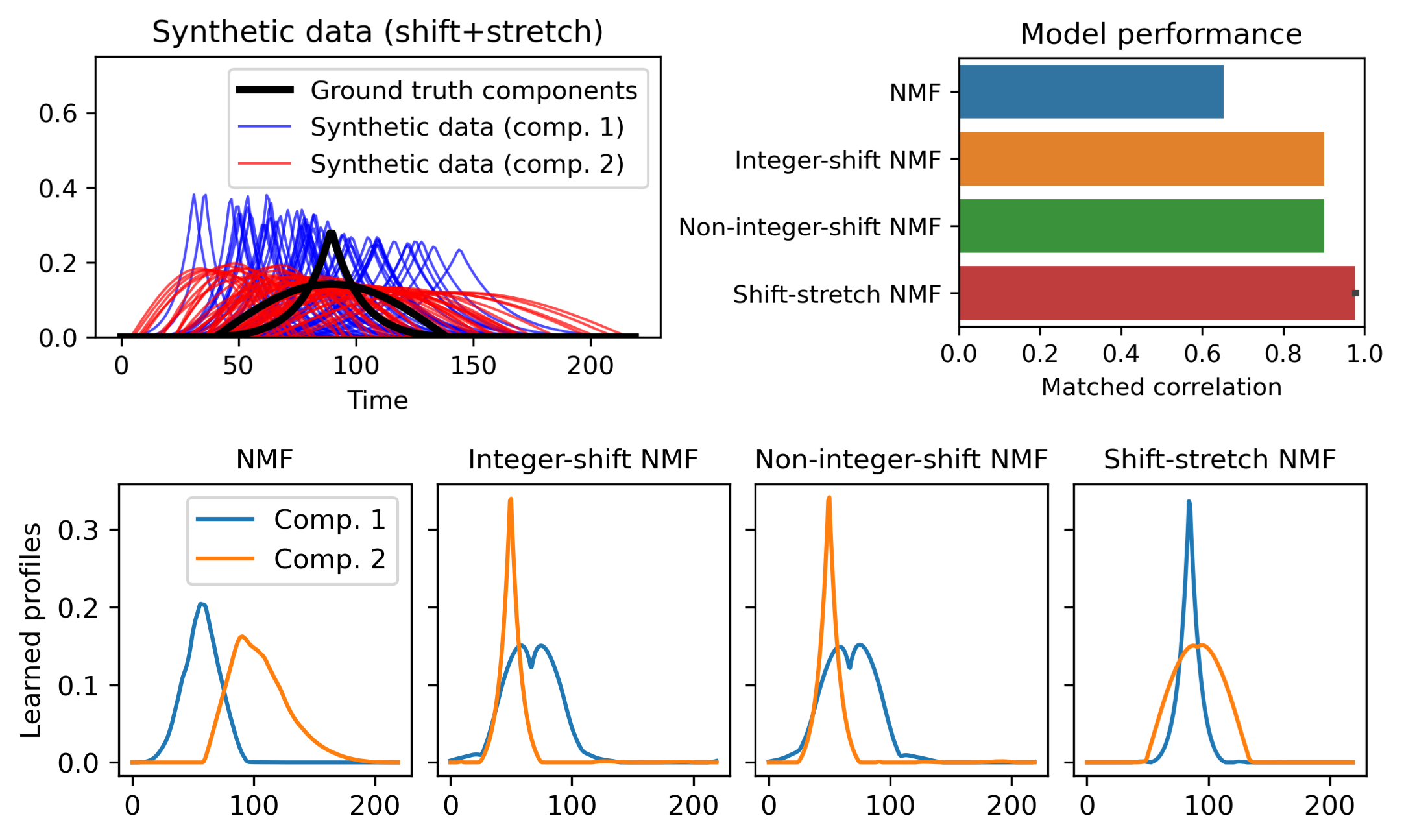}
    \caption{Two-component NMF models applied to two-component synthetic data with random shifts and stretches. Model performance is measured using matched correlation of the $\mathbf A$-matrix to the ground truth design matrix (variance is over 25 repeated model estimates), while the learned profiles (bottom) correspond to the $\mathbf S$-matrix of the estimate with the lowest loss.}
    \label{fig:synthetic}
\end{figure}

\section{Results}\label{sec:results}
\subsection{Synthetic data}

We generated synthetic data by applying random integer shifts and stretches between $\left[-N/4,N/4\right]$ to two true components: One was half of a cosine period (soft peak), and the other was a Laplace probability distribution (sharp hump) (Fig.~\ref{fig:synthetic}). Both ground truth components were pre- and appended by zeros to avoid the applied shifts and stretches to cause data to extend outside the data domain. 100 synthetic channels were generated for each component and model performance measured using matched correlation between the ground-truth and the learned $\mathbf A$. 

The Shift-stretch NMF outperformed the other models in identifying the true labels (Fig.~\ref{fig:synthetic}). The matched correlation coefficients were not precisely equal to one, which we attribute to the frequency domain loss function causing a loss of precision in the learned temporal profiles. The learned profiles were closer to the ground truth for the Shift-stretch model, while the shift models were not able to reconstruct particularly the half-cosine component, which had two humps instead of one. In comparison, the conventional NMF did not yield satisfactory component profiles. 

\subsection{Pig brain SPECT data}

\begin{figure}
    \centering
    \includegraphics[width=0.70\linewidth]{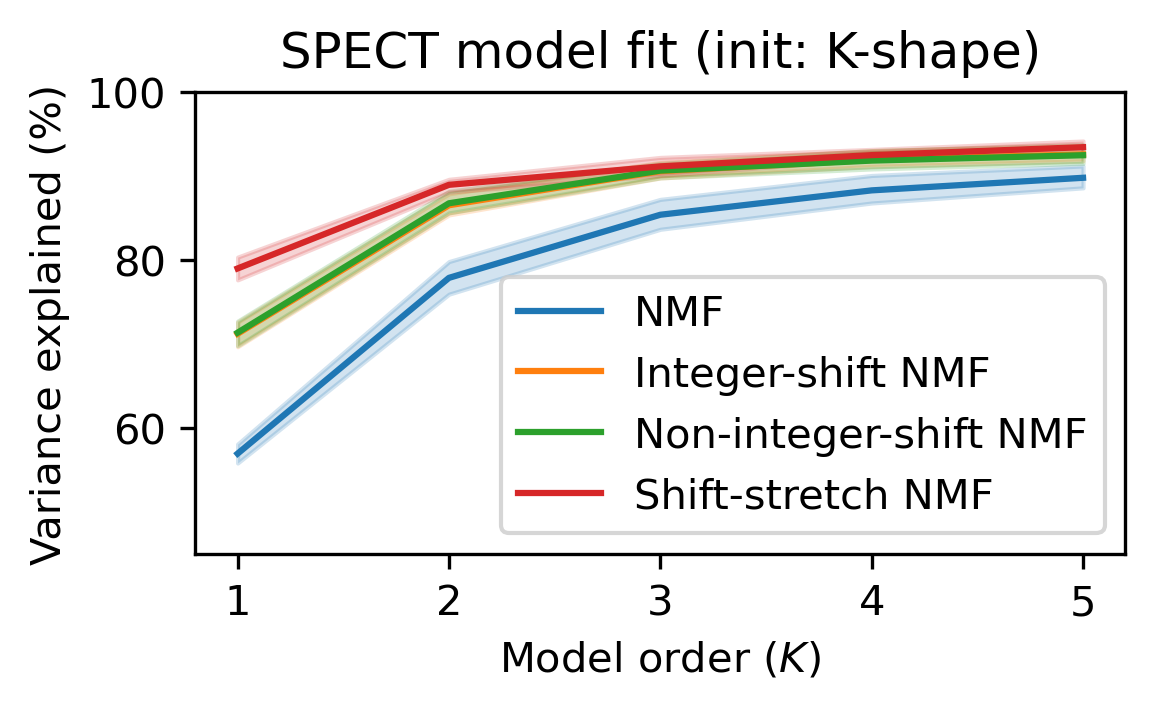}
    \caption{Effect of model order $K$ on variance explained (shaded area is variance over pigs).}
    \label{fig:spect_results}
\end{figure}

Using brain SPECT data from five pigs, we observed that the Shift-stretch NMF model (applied to each pig separately) outperformed the other models across model orders, particularly for $K=1$ and $K=2$, indicating that the added stretching parameters are mostly beneficial for low numbers of components (Fig.~\ref{fig:spect_results}). With more components, Shift-models and even the conventional NMF are able to fit the data comparably to Shift-stretch NMF.


\begin{figure}[h]
    \centering
    \includegraphics[width=0.98\linewidth]{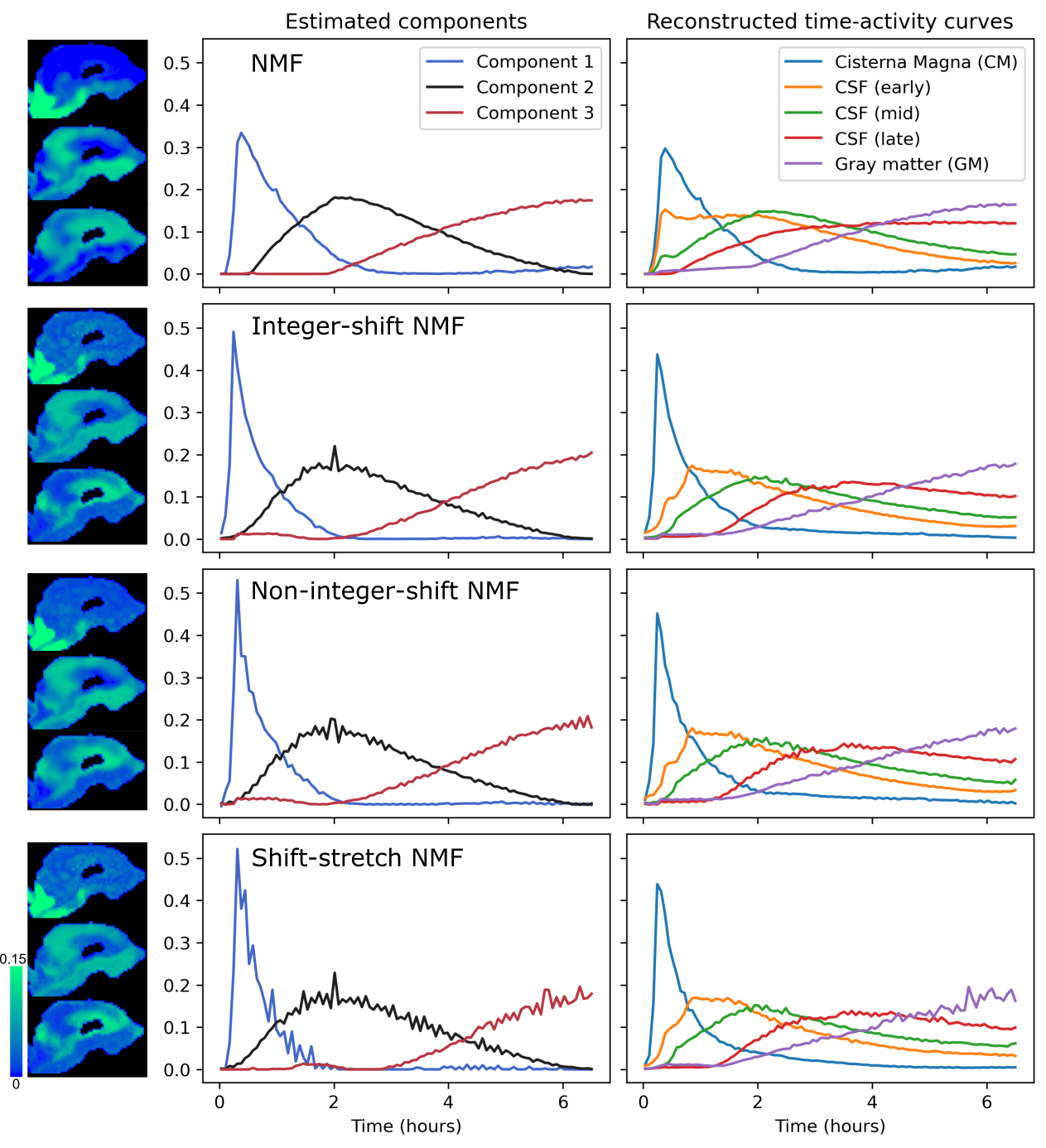}
    \caption{Extracted channel maps (left), temporal profiles (mid), and selected channel reconstructions (right) for pig-5 using $K=3$ components for the four tested models. CSF: cerebrospinal fluid.}
    \label{fig:segmentations}
\end{figure}

Example model fits for $K=3$ components for one pig (pig-5, the same as in Fig.~\ref{fig:spect}) are shown in Fig.~\ref{fig:segmentations}. For all models, one component corresponds roughly to CM - the infusion site, the second one to a broad CSF component, and the third one to gray matter. While the segmentation maps appear similar, the boundaries between clusters appear slightly sharper in NMF and Non-integer-shift NMF, i.e., the models where both $\mathbf A$ and $\mathbf S$ were optimized non-linearly. The temporal profiles were more different between models: The shift and stretch models had a significantly narrower CM component profile, indicating that the conventional NMF averages over many delayed channels. The shift-stretch NMF profiles are more noisy while still contributing to higher explained variance (Fig.~\ref{fig:spect_results}), which we attribute to the manipulation of high-frequency components via zero-padding of spectra (stretching in time domain), such that high-frequency components are non-zero for fewer channels. To avoid this phenomenon, profiles could be reconstructed from only the first $N_{FFT}/2$ frequencies. Fig.~\ref{fig:segmentations}(right) shows the reconstructed example time-activity curves from Fig.~\ref{fig:spect} computed as the right-hand-side of Eqs.~\eqref{eq:nmf}-\eqref{eq:shiftandstrechnmf}. The reconstructions are clearly better for the shift and shift-stretch models. 

\section{Discussion}

We proposed a novel shift- and stretch-invariant version of NMF and applied it to synthetic data as well as SPECT data displaying slow dispersion over multiple tissue compartments (CSF, gray matter). Previous methods for such data did not account for shift and stretch-differences across the brain and would therefore erroneously assume distance-independent data. The method is applicable to data with similar shapes across channels and may therefore also be applied to, e.g., time-locked EEG or physiological measurements. 

The proposed stretching mechanism via truncating/zero-padding in the frequency domain is novel for matrix decompositions, to the best of our knowledge. A previous stretched NMF implementation \cite{gu_stretched_2024} stretched via spline interpolations in the time-domain. We argue that our approach is simpler and has an easier optimization framework. The combination between shifting and stretching results in a versatile matrix decomposition framework, and the implementation in PyTorch means that the framework is very easily altered to related linear decomposition methods such as principal/independent component analysis or sparse coding by replacing non-negativity with orthonormality/independence/sparsity constraints on $\mathbf S$.

Some complications with the model should be mentioned. Shifting is circular, but stretching is not, and the circularity property of phase shifting in the frequency domain is likely seldom desired in real-world scenarios. We propose future studies to approach shifting by constructing a library of candidate temporally shifted profiles, similarly to what we did for stretching. 
Furthermore, the data to which the model is applied should be smooth to avoid large transients, which cause Gibbs ringing when manipulating the frequency components of component profiles - the presented SPECT data did not conform to this. 

In Fig.~\ref{fig:segmentations}, component number three had a small hump in the beginning for most models, which we attribute to ghosting, i.e., the lack of complete uniqueness in the model. Future studies could enforce unimodality by penalizing sign-switching of gradients in $\mathbf S$, which could also remedy the observed noise in component profiles. 

While the components were estimated by shifting and stretching each channel, the estimated temporal profiles for different components cannot be assumed to be aligned to each other (see Fig.~\ref{fig:synthetic}), complicating, e.g., kinetic modeling. In future research, this problem may be approached by selecting a reference region/channel to which other channels are shifted/stretched before computing a weighted average via the estimated $\mathbf A$-matrix. Future modeling could also focus on non-linear optimization of $\mathbf A$ in stretch-invariant models to further refine channel maps.

\pagebreak
\clearpage

\subsection*{Compliance with ethical standards}
All pig experiments were performed in accordance with the European Communities Council Resolves of 22 September 2010 (2010/63/EU) and approved by the Danish Veterinary and Food Administration’s Council for Animal Experimentation (Journal No. 2022–15–2934–00156), and followed the ARRIVE guidelines. 

\subsection*{Funding}
ASO and MLN were supported by JPND research, a Horizon 2020 supported EU joint programme.
JLH and MM were supported by the Independent Research Fund Denmark, grant no. 10.46540/2035-00294B.

\printbibliography

\end{document}